\documentclass{article}

\usepackage{ijcai17}
\usepackage{times}
\usepackage{multirow}
\usepackage[american]{babel}
\usepackage{graphicx}
\graphicspath{{figure/}}
\usepackage{subfig}
\usepackage{amsmath,amssymb} 
\usepackage{pifont}
\usepackage{bm}

\usepackage{microtype}
\usepackage{url}

\begin{document}

\title{Deep Descriptor Transforming for Image Co-Localization\thanks{The first two authors contributed equally to this work. This research was supported by NSFC (61422203, 61333014) and 973 Program (2014CB340501). C. Shen's participation was in part supported by ARC Future Fellowship (FT120100969). X.-S. Wei's contribution was made when visiting The University of Adelaide, and his participation was supported by China Scholarship Council. J. Wu is the corresponding author.}}
\author{Xiu-Shen Wei$^1$, Chen-Lin Zhang$^1$, Yao Li$^2$, Chen-Wei Xie$^1$, Jianxin Wu$^1$, Chunhua Shen$^2$, Zhi-Hua Zhou$^1$\\
$^1$National Key Laboratory for Novel Software Technology, Nanjing University, Nanjing, China\\
$^2$The University of Adelaide, Adelaide, Australia\\
\{weixs,zhangcl,xiecw,wujx,zhouzh\}@lamda.nju.edu.cn, \{yao.li01,chunhua.shen\}@adelaide.edu.au
}

\maketitle

\begin{abstract}
Reusable model design becomes desirable with the rapid expansion of machine learning applications. In this paper, we focus on the reusability of pre-trained deep convolutional models. Specifically, different from treating pre-trained models as feature extractors, we reveal more treasures beneath convolutional layers, i.e., the convolutional activations could act as a detector for the common object in the image co-localization problem. We propose a simple but effective method, named Deep Descriptor Transforming (DDT), for evaluating the correlations of descriptors and then obtaining the category-consistent regions, which can accurately locate the common object in a set of images. Empirical studies validate the effectiveness of the proposed DDT method. On benchmark image co-localization datasets, DDT consistently outperforms existing state-of-the-art methods by a large margin. Moreover, DDT also demonstrates good generalization ability for unseen categories and robustness for dealing with noisy data.
\end{abstract}

\section{Introduction}

Model reuse~\cite{learnware} attempts to construct a model by utilizing existing available models, mostly trained for other tasks, rather than building a model from scratch. Particularly in deep learning, since deep convolutional neural networks have achieved great success in various tasks involving images, videos, texts and more, there are several studies have the flavor of reusing deep models pre-trained on ImageNet~\cite{russaijcv2015}.

In machine learning, the Fixed Model Reuse scheme~\cite{yangaaai2017} is proposed recently for using the sophisticated fixed model/features from a well-trained deep model, rather than transferring with pre-trained weights. In computer vision, pre-trained models on ImageNet have also been successfully adopted to various usages, e.g., as universal feature extractors~\cite{wangiccv2015,yaoeccv2016}, object proposal generators~\cite{Amir15ICCV}, etc. In particular, \cite{scda2016} proposed the SCDA method to utilize pre-trained models for both localizing a single fine-grained object (e.g., birds of different species) in each image and retrieving fine-grained images of the same classes/species in an unsupervised fashion. 

\begin{figure}[t]
 \centering
 \includegraphics[width=0.8\columnwidth, height=20em]{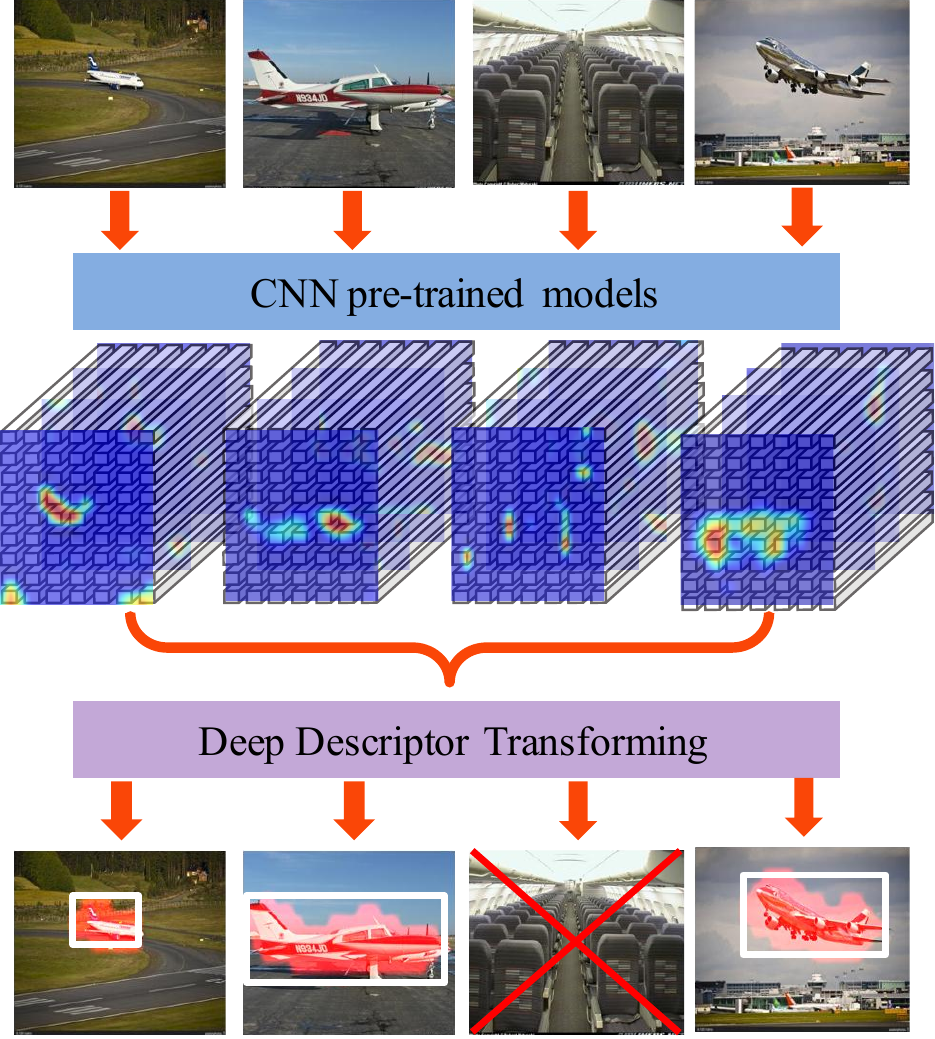}
 \caption{Pipeline of the proposed DDT method for image co-localization. In this instance, the goal is to localize the \emph{airplane} within each image. Note that, there might be few noisy images in the image set. (Best viewed in color.)}
 \label{fig:pipeline}
\vspace{-0.15em}
\end{figure}

In this paper, we reveal that the convolutional activations can be a detector for the \emph{common object} in image co-localization. Image co-localization is a fundamental computer vision problem, which simultaneously localizes objects of the same category across a set of distinct images. Specifically, we propose a simple but effective method named DDT (Deep Descriptor Transforming) for image co-localization. In DDT, the deep convolutional descriptors extracted from pre-trained models are transformed into a new space, where it can evaluate the correlations between these descriptors. By leveraging the correlations among the image set, the common object inside these images can be located automatically without additional supervision signals. The pipeline of DDT is shown in Fig.~\ref{fig:pipeline}. To our best knowledge, this is \emph{the first work} to demonstrate the possibility of convolutional activations/descriptors in pre-trained models \emph{being able to act as a detector for the common object.}

Experimental results show that DDT significantly outperforms existing state-of-the-art methods, including image co-localization and weakly supervised object localization, in both the deep learning and hand-crafted feature scenarios. Besides, we empirically show that DDT has a good generalization ability for unseen images apart from ImageNet. More importantly, the proposed method is robust, because DDT can also detect the noisy images which do not contain the common object. 

\section{Related work}


\subsection{CNN model reuse}

Reusability has been emphasized by~\cite{learnware} as a crucial characteristic of the new concept of \emph{learnware}. It would be ideal if models can be reused in scenarios that are very different from their original training scenarios. Particularly, with the breakthrough in image classification using Convolutional Neural Networks (CNN), pre-trained CNN models trained for one task (e.g., recognition) have also been applied to domains different from their original purposes (e.g., for describing texture or finding object proposals~\cite{Amir15ICCV}). However, for such adaptations of pre-trained models, they still require further annotations in the new domain (e.g., image labels). While, DDT deals with the image co-localization problem in an unsupervised setting.

Coincidentally, several recent works also shed lights on CNN pre-trained model reuse in the unsupervised setting, e.g., SCDA~\cite{scda2016}. SCDA is proposed for handling the fine-grained image retrieval task, where it uses pre-trained models (from ImageNet, which is not fine-grained) to locate main objects in fine-grained images. It is the most related work to ours, even though SCDA is not for image co-localization. Different from our DDT, SCDA assumes only an object of interest in each image, and meanwhile objects from other categories does not exist. Thus, SCDA locates the object using cues from this \emph{single} image assumption. Apparently, it can not work well for images containing diverse objects (cf. Table~\ref{table:voc07} and Table~\ref{table:voc12}), and also can not handle data noise (cf. Sec.~\ref{sec:noise}).

\subsection{Image co-localization}

Image co-localization is a fundamental problem in computer vision, where it needs to discover the common object emerging in only positive sets of example images (without any negative examples or further supervisions). Image co-localization shares some similarities with image co-segmentation~\cite{cosegijcai,gunheeiccv2011,joulinmccvpr2010}. Instead of generating a precise segmentation of the related objects in each image, co-localization methods aim to return a bounding box around the object. Moreover, co-segmentation has a strong assumption that \emph{every} image contains the object of interest, and hence is unable to handle noisy images.

Additionally, co-localization is also related to weakly supervised object localization (WSOL)~\cite{wsolijcai,bilencvpr2015,wangeccv2014,sivacvpr2011}. But the key difference between them is WSOL requires manually-labeled negative images whereas co-localization does not. Thus, WSOL methods could achieve better localization performance than co-localization methods. However, our DDT performs comparably with state-of-the-art WSOL methods and even outperforms them (cf. Table~\ref{table:voc07weak}).

Recently, there are also several co-localization methods based on pre-trained models, e.g.,~\cite{yaoeccv2016,wangeccv2014}. But, these methods just treated pre-trained models as simple feature extractors to extract the fully connected representations, which did not leverage the original correlations between deep descriptors among convolutional layers. Moreover, these methods also needed object proposals as a part of their object discovery, which made them highly dependent on the quality of object proposals. In addition, almost all the previous co-localization methods can not handle noisy data, except for~\cite{tangcvpr2014}.

Comparing with previous works, our DDT is unsupervised, without utilizing bounding boxes, additional image labels or redundant object proposals. Images only need one forward run through a pre-trained model. Then, efficient deep descriptor transforming is employed for obtaining the category-consistent image regions. DDT is very easy to implement, and surprisingly has good generalization ability and robustness.

\section{The proposed method}


\subsection{Preliminary}

The following notations are used in the rest of this paper. The term ``feature map'' indicates the convolution results of one channel; the term ``activations'' indicates feature maps of all channels in a convolution layer; and the term ``descriptor'' indicates the $d$-dimensional component vector of activations.

Given an input image $I$ of size $H\times W$, the activations of a convolution layer are formulated as an order-3 tensor $T$ with $h\times w\times d$ elements. $T$ can be considered as having $h\times w$ cells and each cell contains one $d$-dimensional deep descriptor. For the $n$-th image, we denote its corresponding deep descriptors as $X^n=\left \{\bm{x}^n_{\left(i,j\right)}\in \mathcal{R}^{d}\right \}$, where $\left(i,j \right)$ is a particular cell ($i\in \left \{1,\ldots,h\right \}, j\in \left \{1,\ldots,w \right \}$) and $n\in \left \{1,\ldots,N\right \}$.

\subsection{SCDA recap}\label{sec:scda}

Since SCDA~\cite{scda2016} is the most related work to ours, we hereby present a recap of this method. SCDA is proposed for dealing with the fine-grained image retrieval problem. It employs pre-trained models to select the meaningful deep descriptors by localizing the main object in fine-grained images unsupervisedly. In SCDA, it assumes that each image contains only one main object of interest and without other categories' objects. Thus, the object localization strategy is based on the activation tensor of a \emph{single} image.

Concretely, for an image, the activation tensor is added up through the depth direction. Thus, the $h\times w\times d$ 3-D tensor becomes a $h\times w$ 2-D matrix, which is called the ``aggregation map'' in SCDA. Then, the mean value $\bar{a}$ of the aggregation map is regarded as the threshold for localizing the object. If the activation response in the position $\left(i,j\right)$ of the aggregation map is larger than $\bar{a}$, it indicates the object might appear in that position.

\subsection{Deep descriptor transforming (DDT)}

What distinguishes DDT from SCDA is that we can leverage the correlations beneath the whole \emph{image set}, instead of a \emph{single} image. Additionally, different from weakly supervised object localization, we do not have either image labels or negative image sets in WSOL, so that the information we can use is only from the pre-trained models. Here, we transform the deep descriptors in convolutional layers to mine the hidden information for co-localizing common objects.

Principal component analysis (PCA)~\cite{pca1901} is a statistical procedure, which uses an orthogonal transformation to convert a set of observations of possibly correlated variables into a set of linearly uncorrelated variables (i.e., the principal components). This transformation is defined in such a way that the first principal component has the largest possible variance, and each succeeding component in turn has the highest variance possible under the constraint that it is orthogonal to all the preceding components.

\begin{figure*}[t]
 \centering
 \includegraphics[width=0.99\textwidth, height=17.5em]{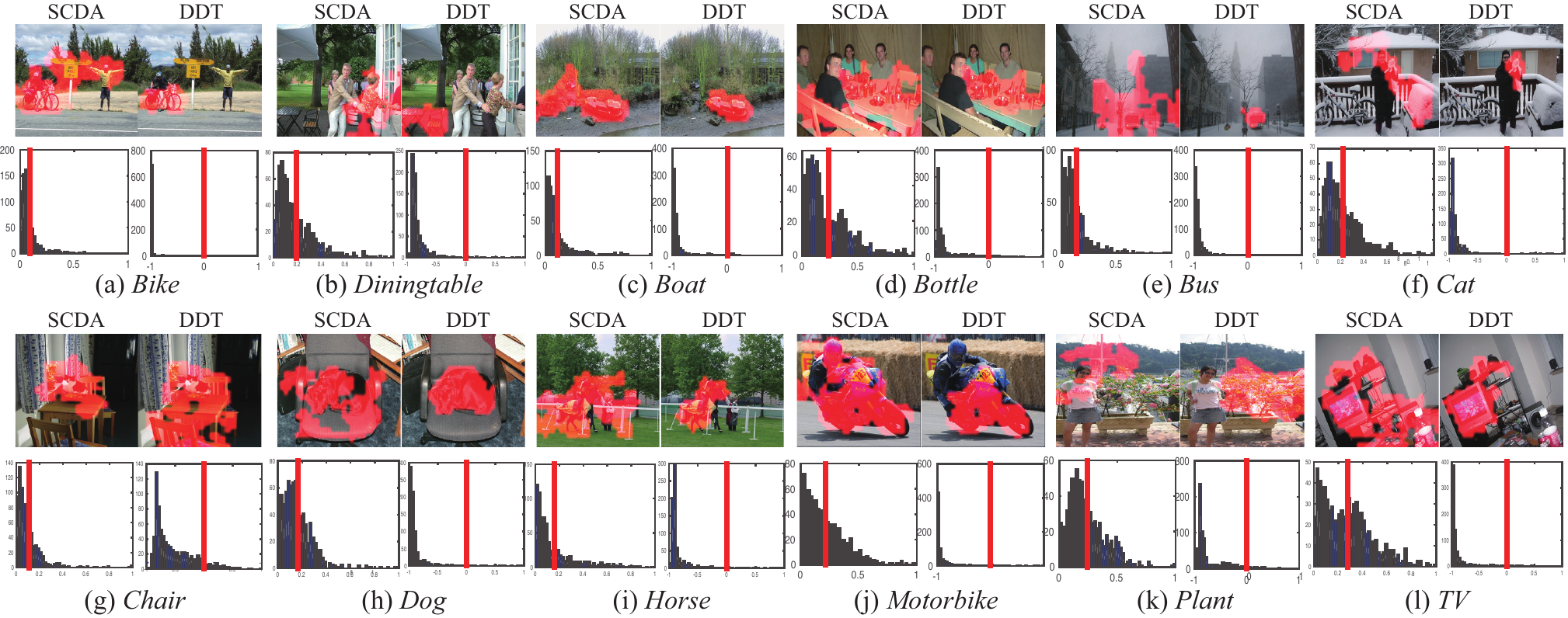}
\vspace{-0.5em}
 \caption{Examples from twelve randomly sampled classes of \emph{VOC 2007}. The first column of each subfigure are produced by SCDA, the second column are by our DDT. The red vertical lines in the histogram plots indicate the corresponding thresholds for localizing objects. The selected regions in images are highlighted in red. (Best viewed in color and zoomed in.)}
 \label{fig:heatmap2}
\vspace{-0.3em}
\end{figure*}

PCA is widely used in machine learning and computer vision for dimension reduction~\cite{moijcai2013,quanijcai2011,tianhaotkde2009,ianijcai2009}, noise reduction~\cite{liangpeitgars,feipingijcai2011} and so on. Specifically, in this paper, we utilize PCA as projection directions for transforming these deep descriptors $\{\bm{x}_{\left(i,j\right)}\}$ to evaluate their correlations. Then, on each projection direction, the corresponding principal component's values are treated as the cues for image co-localization, especially the first principal component. Thanks to the property of this kind of transforming, DDT is also able to handle data noise.

In DDT, for a set of $N$ images containing objects from the same category, we first collect the corresponding convolutional descriptors ($X^1,\ldots,X^N$) by feeding them into a pre-trained CNN model. Then, the mean vector of all the descriptors is calculated by:
\begin{equation}
\label{eq:mean}
\bar{\bm{x}} = \frac{1}{K} \sum_n \sum_{i,j} \bm{x}_{\left(i,j\right)}^n \,,
\end{equation}
where $K=h\times w\times N$. Note that, here we assume each image has the same number of deep descriptors (i.e., $h\times w$) for presentation clarity. Our proposed method, however, can handle input images with arbitrary resolutions.

Then, after obtaining the covariance matrix:
\begin{equation}
\label{eq:cov}
{\rm{Cov}} (\bm{x})=\frac{1}{K} \sum_n \sum_{i,j} (\bm{x}_{\left(i,j\right)}^n-\bar{\bm{x}}) (\bm{x}_{\left(i,j\right)}^n-\bar{\bm{x}})^\top \,,
\end{equation}
we can get the eigenvectors $\bm{\xi}_1,\ldots, \bm{\xi}_d$ of ${\rm Cov}(\bm{x})$ which correspond to the sorted eigenvalues $\lambda_1\geq \cdots \geq \lambda_d \geq 0$.

As aforementioned, since the first principal component has the largest variance, we take the eigenvector $\bm{\xi}_1$ corresponding to the largest eigenvalue as the main projection direction. For the deep descriptor at a particular position $\left(i,j\right)$ of an image, its first principal component $p^1$ is calculated as follows:
\begin{equation}
\label{eq:firstpc}
p_{(i,j)}^1 = \bm{\xi}^\top_1 \left(\bm{x}_{\left(i,j\right)} - \bar{\bm{x}} \right) \,.
\end{equation}
According to their spatial locations, all $p_{(i,j)}^1$ from an image are combined into a 2-D matrix whose dimensions are $h\times w$. We call that matrix as \emph{indicator matrix}:
\begin{equation}
\label{eq:indmat}
P^1 = \left[
  			\begin{array}{cccc}  
          					p_{(1,1)}^1 & p_{(1,2)}^1 & \ldots & p_{(1,w)}^1 \\  
						p_{(2,1)}^1 & p_{(2,2)}^1 & \ldots & p_{(2,w)}^1 \\
          					\vdots & \vdots & \ddots & \vdots \\  
          					p_{(h,1)}^1 & p_{(h,2)}^1 & \ldots & p_{(h,w)}^1
 			\end{array}  
 	      \right]\,.
\end{equation}

$P^1$ contains positive (negative) values which can reflect the positive (negative) correlations of these deep descriptors. The larger the absolute value is, the higher the positive (negative) correlation will be. Because $\bm{\xi}_1$ is obtained through all $N$ images, the positive correlation could indicate the \emph{common characteristic} through $N$ images. Specifically, in the image co-localization scenario, the corresponding positive correlation indicates indeed the \emph{common object} inside these images.

Therefore, the value zero could be used as a natural threshold for dividing $P^1$ of one image into two parts: one part has positive values indicating the common object, and the other part has negative values presenting background objects rarely appear. In addition, if $P^1$ of an image has no positive value, it indicates that no common object exists in that image, which can be used for detecting noisy images. In practice, $P^1$ is resized by the nearest interpolation, such that its size is the same as that of the input image. Meanwhile, we collect the largest connected component of the positive regions of $P^1$ (as what is done in~\cite{scda2016}). Based on these positive correlation values and the zero threshold, the minimum rectangle bounding box which contains the largest connected component of positive regions is returned as our object co-localization prediction.

\subsection{Discussions and analyses}

In this section, we investigate the effectiveness of DDT by comparing with SCDA.

As shown in Fig.~\ref{fig:heatmap2}, the object localization regions of SCDA and DDT are highlighted in red. Because SCDA only considers the information from a single image, in Fig.~\ref{fig:heatmap2} (a),  ``bike'', ``person'' and even ``guide-board'' are all detected as main objects. Furthermore, we normalize the values (all positive) of the aggregation map of SCDA into the scale of $\left[0,1\right]$, and calculate the mean value (which is taken as the object localization threshold in SCDA). The histogram of the normalized values in aggregation map is also shown in that figure. The red vertical line corresponds to the threshold. We can find that, beyond the threshold, there are still many values. It gives an explanation about why SCDA highlights more regions.

Whilst, for DDT, it leverages the whole image set to transform these deep descriptors into $P^1$. Thus, for the \emph{bicycle} class, DDT can accurately locate the ``bicycle'' object. The histogram is also drawn. But, $P^1$ has both positive and negative values. We normalize $P^1$ into the $\left[-1,1\right]$ scale this time. Apparently, few values are larger than the DDT threshold (i.e.,~0). More importantly, many values are close to $-1$ which indicates the strong negative correlation. This observation validates the effectiveness of DDT in image co-localization. As another example shown in Fig.~\ref{fig:heatmap2} (b), SCDA even wrongly locates ``person'' in the image belonging to the \emph{diningtable} class. While, DDT can correctly and accurately locate the ``diningtable'' image region. In Fig.~\ref{fig:heatmap2}, more examples are presented. In that figure, some failure cases can be also found, e.g., the \emph{chair} class in Fig.~\ref{fig:heatmap2} (g).

In addition, the normalized $P^1$ can be also used as localization probability scores. Combining it with conditional random filed techniques might produce more accurate object boundaries. Thus, DDT can be modified slightly in that way, and then perform the co-segmentation problem. More importantly, different from other co-segmentation methods, DDT can detect noisy images while other methods can not.

\section{Experiments}

In this section, we first introduce the evaluation metric and datasets used in image co-localization. Then, we compare the empirical results of our DDT with other state-of-the-arts on these datasets. The computational cost of DDT is reported too. Moreover, the results in Sec.~\ref{sec:unseen} and Sec.~\ref{sec:noise} illustrate the generalization ability and robustness of the proposed method. Finally, our further study in Sec.~\ref{sec:future} reveals DDT might deal with part-based image co-localization, which is a novel and challenging problem.

In our experiments, the images keep the original image resolutions. For the pre-trained deep model, the publicly available VGG-19 model~\cite{vgg16} is employed to extract deep convolution descriptors from the last convolution layer (before ${\rm pool}_5$). We use the open-source library MatConvNet~\cite{matconvnet} for conducting experiments. All the experiments are run on a computer with Intel Xeon E5-2660 v3, 500G main memory, and a K80 GPU.

\subsection{Evaluation metric and datasets}

Following previous image co-localization works~\cite{yaoeccv2016,chicvpr2015,tangcvpr2014}, we take the correct localization (CorLoc) metric for evaluating the proposed method. CorLoc is defined as the percentage of images correctly localized according to the PASCAL-criterion~\cite{voc2015}: $\frac{{\rm area} (B_{\rm p} \cap B_{\rm gt})}{{\rm area} (B_{\rm p} \cup B_{\rm gt})}>0.5$, where $B_{\rm p}$ is the predicted bounding box and $B_{\rm gt}$ is the ground-truth bounding box. All CorLoc results are reported in percentages.

Our experiments are conducted on four challenging datasets commonly used in image co-localization, i.e., the \emph{Object Discovery} dataset~\cite{rubinscvpr2013}, the \emph{PASCAL VOC 2007} / \emph{VOC 2012} dataset~\cite{voc2015} and the \emph{ImageNet Subsets}~\cite{yaoeccv2016}.

\begin{table}[t!]
 \caption{Comparisons of CorLoc on \emph{Object Discovery}.} \label{table:objdisc}
 \centering
 \vspace{-0.7em}
 \scriptsize
 \begin{tabular}{|c||c|c|c||c|}
  \hline
  {Methods}  & {\emph{Airplane}} & {\emph{Car}} & {\emph{Horse}} & {\textbf{Mean}} \\  
  \hline
  \cite{joulincoscvpr2010}  & 32.93  & 66.29 &  54.84 &  51.35 \\
  \cite{joulinmccvpr2010}  & 57.32  & 64.04 &  52.69  & 58.02\\
  \cite{rubinscvpr2013}  & 74.39  & 87.64  & 63.44 &  75.16\\
  \cite{tangcvpr2014}  & 71.95 &  93.26  & 64.52  & 76.58\\
  SCDA & 87.80  &  	86.52  &  	75.37  &  	83.20 \\
  \cite{chicvpr2015}  & 82.93  & 94.38  & 75.27 &  84.19\\
  \hline
  Our DDT & \textbf{91.46}	 & \textbf{95.51}	 & \textbf{77.42}	 & \textbf{88.13}\\
  \hline
 \end{tabular}
\vspace{-1em}
\end{table}

For experiments on the \emph{VOC} datasets, we follow~\cite{chicvpr2015,yaoeccv2016,joulineccv2014} to use all images in the \emph{trainval} set (excluding images that only contain object instances annotated as \emph{difficult} or \emph{truncated}). For \emph{Object Discovery}, we use the 100-image subset following~\cite{rubinscvpr2013,chicvpr2015} in order to make an appropriate comparison with other methods.

In addition, \emph{Object Discovery} has 18\%, 11\% and 7\% noisy images in the \emph{Airplane}, \emph{Car} and \emph{Horse} categories, respectively. These noisy images contain no object belonging to their category, as the third image shown in Fig.~\ref{fig:pipeline}. Particularly, in Sec.~\ref{sec:noise}, we quantitatively measure the ability of our proposed DDT to identify these noisy images.

To further investigate the generalization ability of DDT, \emph{ImageNet Subsets}~\cite{yaoeccv2016} are used, which contain six subsets/categories. These subsets are held-out categories from the 1000-label ILSVRC classification~\cite{russaijcv2015}. That is to say, these subsets are ``unseen'' by pre-trained CNN models. Experimental results in Sec.~\ref{sec:unseen} show that DDT is insensitive to the object category.

\begin{table*}[th!]
 \caption{Comparisons of the CorLoc metric with state-of-the-art co-localization methods on \emph{VOC 2007}.} \label{table:voc07}
 \centering
 \vspace{-0.7em}
 \setlength{\tabcolsep}{3.5pt}
 \scriptsize
 \begin{tabular}{|c||c|c|c|c|c|c|c|c|c|c|c|c|c|c|c|c|c|c|c|c||c|}
  \hline
  {Methods} & \emph{aero} & \emph{bike} & \emph{bird}& \emph{boat}& \emph{bottle}& \emph{bus}& \emph{car}& \emph{cat}& \emph{chair}& \emph{cow}& \emph{table}& \emph{dog}& \emph{horse}& \emph{mbike}& \emph{person}& \emph{plant}& \emph{sheep}& \emph{sofa}& \emph{train}& \emph{tv} & \textbf{Mean}\\  
  \hline
  \cite{joulineccv2014}  & 32.8  & 17.3  & 20.9 &  18.2  & 4.5  & 26.9 &  32.7  & 41.0 &  5.8  & 29.1  & \textbf{34.5}  & 31.6  & 26.1  & 40.4  & 17.9  & 11.8 &  25.0  & 27.5  & 35.6  & 12.1  & 24.6  \\
  SCDA & 54.4   & 	27.2 	  & 43.4 	  & 13.5   & 	2.8   & 	39.3 	  & 44.5   & 	48.0   & 	6.2 	  & 32.0   & 	16.3   & 	49.8 	  & 51.5   & 	49.7   & 	7.7   & 	6.1  &  	22.1  &  	22.6 	  & 46.4   & 	6.1   & 	29.5 \\
  \cite{chicvpr2015}   & 50.3 &  42.8 &  30.0  & 18.5  & 4.0  & 62.3  & \textbf{64.5} &  42.5  & 8.6  & \textbf{49.0}  & 12.2  & 44.0  & 64.1  & 57.2  & 15.3  & 9.4  & 30.9  & 34.0 &  61.6  & \textbf{31.5}  & 36.6\\
  \cite{yaoeccv2016}   & \textbf{73.1}  & 45.0 &  43.4  & \textbf{27.7} &  6.8  & 53.3  & 58.3  & 45.0  & 6.2  & 48.0  & 14.3  & 47.3  & 69.4  & 66.8 &  \textbf{24.3}  & 12.8  & \textbf{51.5}  & 25.5  & 65.2  & 16.8  & 40.0 \\
  \hline
  Our DDT  & 67.3 	& \textbf{63.3}  & 	\textbf{61.3}  & 	22.7  & 	\textbf{8.5} 	 & \textbf{64.8}  & 	57.0  & 	\textbf{80.5} 	 & \textbf{9.4} 	 & \textbf{49.0} 	 & 22.5 	 & \textbf{72.6} &  	\textbf{73.8}  & 	\textbf{69.0}  & 	7.2  & 	\textbf{15.0}  & 	35.3  & 	\textbf{54.7} 	 & \textbf{75.0} 	 & 29.4 	 & \textbf{46.9} \\
  \hline
 \end{tabular}
\end{table*}

\begin{table*}[th!]
 \caption{Comparisons of the CorLoc metric with state-of-the-art co-localization methods on \emph{VOC 2012}.} \label{table:voc12}
 \centering
 \vspace{-0.7em}
 \setlength{\tabcolsep}{3.6pt}
 \scriptsize
 \begin{tabular}{|c||c|c|c|c|c|c|c|c|c|c|c|c|c|c|c|c|c|c|c|c||c|}
  \hline
  {Methods} & \emph{aero} & \emph{bike} & \emph{bird}& \emph{boat}& \emph{bottle}& \emph{bus}& \emph{car}& \emph{cat}& \emph{chair}& \emph{cow}& \emph{table}& \emph{dog}& \emph{horse}& \emph{mbike}& \emph{person}& \emph{plant}& \emph{sheep}& \emph{sofa}& \emph{train}& \emph{tv} & \textbf{Mean}\\  
  \hline
  SCDA & 60.8  & 	41.7 	 & 38.6 	 & 21.8  & 	7.4  & 	67.6  & 	38.8 	 & 57.4  & 	16.0  & 	34.0  &  	23.9  & 	53.8  & 	47.3  & 	54.8 &  	7.9  & 	9.9  & 	25.3  & 	23.2  & 	50.2  & 	10.1  & 	34.5  \\
  \cite{chicvpr2015}  & 57.0 	 & 41.2  & 	36.0  & 	26.9  & 	5.0 	 & 81.1 	 & \textbf{54.6}  & 	50.9  & 	18.2  & 	54.0  & 	\textbf{31.2}  & 	44.9  & 	61.8  & 	48.0 &  	13.0 	 & 11.7  & 	51.4  & 	45.3  & 	64.6  & 	\textbf{39.2}  & 	41.8   \\
  \cite{yaoeccv2016}   &65.7 	 & 57.8  & 	47.9 	 & 28.9 	 & 6.0 	 & 74.9 	 & 48.4  & 	48.4  & 	14.6  & 	\textbf{54.4}  & 	23.9  & 	50.2  & 	\textbf{69.9}  & 	68.4  & 	\textbf{24.0}  & 	14.2  & 	\textbf{52.7} &  	30.9  & 	72.4 	 & 21.6 	 & 43.8  \\
  \hline
  Our DDT  & \textbf{76.7}  & 	\textbf{67.1} 	 & \textbf{57.9} 	 & \textbf{30.5}  & 	\textbf{13.0}  & 	\textbf{81.9}  & 	48.3  & 	\textbf{75.7} 	 & \textbf{18.4}  & 	48.8  & 	27.5 	 & \textbf{71.8}  & 	66.8 &  	\textbf{73.7} 	 & 6.1  & 	\textbf{18.5}  & 	38.0  & 	\textbf{54.7}  & 	\textbf{78.6}  & 	34.6 	 & \textbf{49.4}  \\
  \hline
 \end{tabular}
\end{table*}

\subsection{Comparisons with state-of-the-arts}


\subsubsection{Comparisons to image co-localization methods}\label{sec:comparison}

We first compare the results of DDT to state-of-the-arts (including SCDA) on \emph{Object Discovery} in Table~\ref{table:objdisc}. For SCDA, we also use VGG-19 to extract the convolution descriptors and perform experiments. As shown in that table, DDT outperforms other methods by about 4\% in the mean CorLoc metric. Especially for the \emph{airplane} class, it is about 10\% higher than that of~\cite{chicvpr2015}. In addition, note that the images of each category in this dataset contain only one object, thus, SCDA can perform well.

For \emph{VOC 2007} and \emph{2012}, these datasets contain diverse objects per image, which is more challenging than \emph{Object Discovery}. The comparisons of the CorLoc metric on these two datasets are reported in Table~\ref{table:voc07} and Table~\ref{table:voc12}, respectively. It is clear that on average our DDT outperforms the previous state-of-the-arts (based on deep learning) by a large margin on both two datasets. Moreover, DDT works well on localizing small common objects, e.g., ``bottle'' and ``chair''. In addition, because most images of these datasets have multiple objects, which do not obey  SCDA's assumption, SCDA performs badly in the complicated environment. For fair comparisons, we also use VGG-19 to extract the fully connected representations of the object proposals in~\cite{yaoeccv2016}, and then perform the remaining processes of their method (the source codes are provided by the authors). As aforementioned, due to the high dependence on the quality of object proposals, their mean CorLoc metric of VGG-19 is 41.9\% and 45.6\% on \emph{VOC 2007} and \emph{2012}, respectively. The improvements are limited, and the performance is still significantly worse than ours.

\subsubsection{Comparisons to weakly supervised localization methods}

To further verify the effectiveness of DDT, we also compare it with some state-of-the-art methods for weakly supervised object localization. Table~\ref{table:voc07weak} illustrates these empirical results on \emph{VOC 2007}. Particularly, DDT achieves 46.9\% on average which is higher than most WSOL methods in the literature. But, it still has a small gap (0.8\% lower) with that of \cite{wangeccv2014} which is also a deep learning based approach. This is understandable as we do \emph{not} use any negative data for co-localization. Meanwhile, our DDT can easily extend to handle negative data and thus perform WSOL. Moreover, DDT could handle noisy data (cf. Sec.~\ref{sec:noise}). But, existing WSOL methods are not designed to deal with noise.

\begin{table*}[th!]
 \caption{Comparisons of the CorLoc metric with weakly supervised object localization methods on \emph{VOC 2007}. Note that, the ``$\checkmark$'' in the ``Neg.'' column indicates that these WSOL methods require access to a negative image set, whereas our DDT does not.} \label{table:voc07weak}
 \centering
 \vspace{-0.7em}
 \setlength{\tabcolsep}{3pt}
 \scriptsize
 \begin{tabular}{|c|c||c|c|c|c|c|c|c|c|c|c|c|c|c|c|c|c|c|c|c|c||c|}
  \hline
  {Methods}  & Neg. & \emph{aero} & \emph{bike} & \emph{bird}& \emph{boat}& \emph{bottle}& \emph{bus}& \emph{car}& \emph{cat}& \emph{chair}& \emph{cow}& \emph{table}& \emph{dog}& \emph{horse}& \emph{mbike}& \emph{person}& \emph{plant}& \emph{sheep}& \emph{sofa}& \emph{train}& \emph{tv} & \textbf{Mean}\\  
  \hline
  \cite{shiiccv2013} &$\checkmark$ & 67.3  & 	54.4 	 & 34.3  & 	17.8  & 	1.3 	 & 46.6 	 & 60.7 	 & 68.9  & 	2.5 	 & 32.4 	 & 16.2 	 & 58.9 	 & 51.5  & 	64.6  & 	18.2 	 & 3.1  & 	20.9 	 & 34.7 	 & 63.4 	 & 5.9  & 	36.2  \\
  \cite{cinbiscvpr2014} &$\checkmark$ & 56.6 	 & 58.3 &  	28.4  & 	20.7  & 	6.8 	 & 54.9  & 	69.1 &  	20.8 	 & 9.2  & 	50.5 	 & 10.2  & 	29.0 &  	58.0 	 & 64.9  & 	36.7  & 	18.7  & 	56.5  & 	13.2  & 	54.9  & 	59.4 	 & 38.8  \\
  \cite{wangiccv2015} & $\checkmark$& 37.7 	 & 58.8 &  	39.0 &  	4.7  & 	4.0 	 & 48.4 	 & 70.0 	 & 63.7 	 & 9.0  & 	54.2  & 	\textbf{33.3} 	 & 37.4 	 & 61.6  & 	57.6 &  	30.1 	 & 31.7  & 	32.4 &  	52.8  & 	49.0  & 	27.8 	 & 40.2 \\
  \cite{bilencvpr2015} &$\checkmark$ & 66.4  & 	59.3  & 	42.7 	 & 20.4 	 & \textbf{21.3} 	 & 63.4 	 & \textbf{74.3} 	 & 59.6  & 	21.1  & 	58.2 	 & 14.0 	 & 38.5 	 & 49.5  & 	60.0  & 	19.8 	 & 39.2  & 	41.7 &  	30.1 	 & 50.2  & 	44.1  & 	43.7  \\
  \cite{renpami2016} &$\checkmark$ & 79.2 	 & 56.9  & 	46.0  & 	12.2  & 	15.7  & 	58.4  & 	71.4  & 	48.6  & 	7.2 	 & \textbf{69.9}  & 	16.7 &  	47.4  & 	44.2  & 	\textbf{75.5} 	 & \textbf{41.2} 	 & \textbf{39.6}  & 	47.4  & 	32.2  & 	49.8  & 	18.6  & 	43.9 \\
 \cite{wangeccv2014} &$\checkmark$ & \textbf{80.1} 	 & \textbf{63.9} 	 & 51.5 &  	14.9  & 	21.0  & 	55.7  & 	74.2 &  	43.5  & 	\textbf{26.2}  & 	53.4  & 	16.3  & 	56.7  & 	58.3 	 & 69.5 &  	14.1 	 & 38.3  & 	\textbf{58.8}  & 	47.2  & 	49.1  & 	\textbf{60.9}  & 	\textbf{47.7} \\
  \hline
  Our DDT &  & 67.3 	& 63.3  & 	\textbf{61.3}  & 	\textbf{22.7}  & 	8.5 	 & \textbf{64.8}  & 	57.0  & 	\textbf{80.5} 	 & 9.4 	 & 49.0 	 & 22.5 	 & \textbf{72.6} &  	\textbf{73.8}  & 	69.0  & 	7.2  & 	15.0  & 	35.3  & 	\textbf{54.7} 	 & \textbf{75.0} 	 & 29.4 	 & 46.9 \\
  \hline
 \end{tabular}
\end{table*}

\subsection{Computational costs of DDT}

Here, we take the total 171 images in the \emph{aeroplane} category of \emph{VOC 2007} as examples to report the computational costs. The average image resolution of the 171 images is $350\times 498$. The computational time of DDT has two main components: one is for feature extraction, the other is for deep descriptor transforming. Because we just need the first principal component, the transforming time on all the 120,941 descriptors of 512-d is only 5.7 seconds. The average descriptor extraction time is 0.18 second/image on GPU and 0.86 second/image on CPU, respectively. That shows the efficiency of the proposed DDT method in real-world applications.

\subsection{Unseen classes apart from ImageNet}\label{sec:unseen}

In order to justify the generalization ability of DDT, we also conduct experiments on some images (of six subsets) disjoint with the images from ImageNet. Note that, the six categories of these images are unseen by pre-trained models. The six subsets were provided in~\cite{yaoeccv2016}. Table~\ref{table:unseen} presents the CorLoc metric on these subsets. Our DDT (69.1\% on average) still significantly outperforms other methods on all categories, especially for some difficult objects categories, e.g., \emph{rake} and \emph{wheelchair}. In addition, the mean CorLoc metric of~\cite{yaoeccv2016} based on VGG-19 is 51.6\% on this dataset.

\begin{table}[th!]
 \caption{Comparisons of on image sets disjoint with ImageNet.} \label{table:unseen}
 \centering
 \vspace{-0.7em}
 \setlength{\tabcolsep}{2.8pt}
 \scriptsize
 \begin{tabular}{|c||c|c|c|c|c|c||c|}
  \hline
  {Methods}  & {\emph{Chipmunk}} & {\emph{Rhino}} & {\emph{Stoat}} & {\emph{Racoon}} & \emph{Rake} & \emph{Wheelchair} & \textbf{Mean} \\  
  \hline
  \cite{chicvpr2015}  & 26.6 & 81.8 & 44.2 & 30.1 & 8.3 & 35.3& 37.7 \\
  SCDA & 32.3  & 	71.6  & 	52.9  & 	34.0 	 & 7.6  & 	28.3 	 & 37.8 \\
  \cite{yaoeccv2016} & 44.9 & 81.8& 67.3& 41.8& 14.5& 39.3  & 48.3\\
  \hline
  Our DDT & \textbf{70.3}  & 	\textbf{93.2}  & 	\textbf{80.8} 	 & \textbf{71.8}  & 	\textbf{30.3}  & 	\textbf{68.2} & 	\textbf{69.1} \\
  \hline
 \end{tabular}
 \vspace{-1em}
\end{table}

\begin{figure*}[t]
 \centering
\vspace{-1em}
 \subfloat[\emph{Chipmunk}]  { \includegraphics[width=0.65\columnwidth]{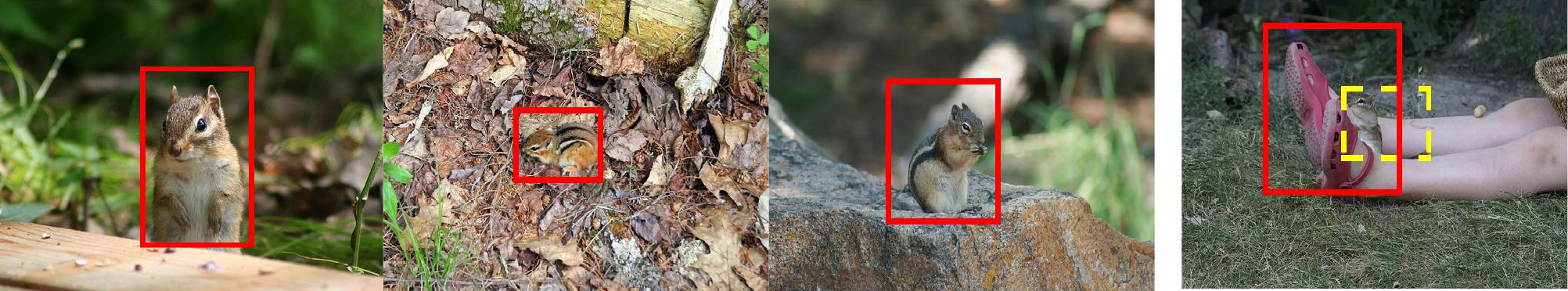} \label{fig:im1} }
	~~
 \subfloat[\emph{Rhino}] { \includegraphics[width=0.65\columnwidth]{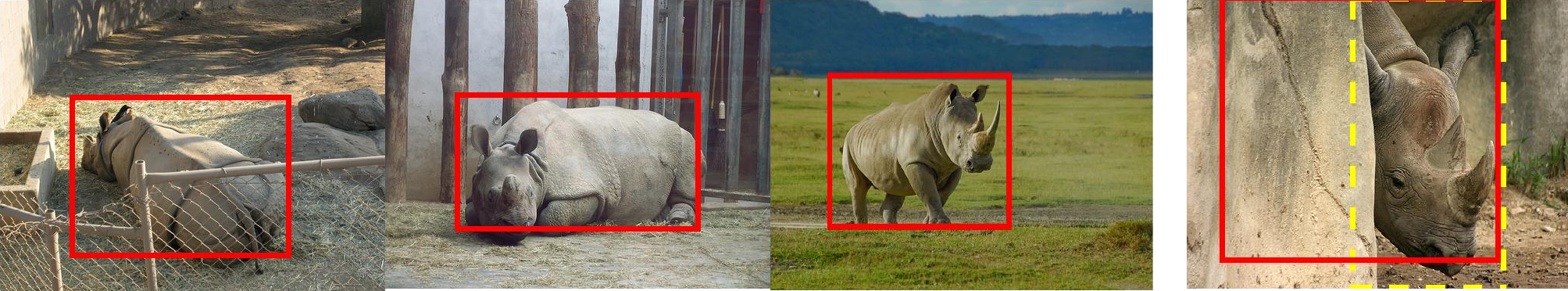} \label{fig:im2} }
	~~
 \subfloat[\emph{Stoat}] { \includegraphics[width=0.65\columnwidth]{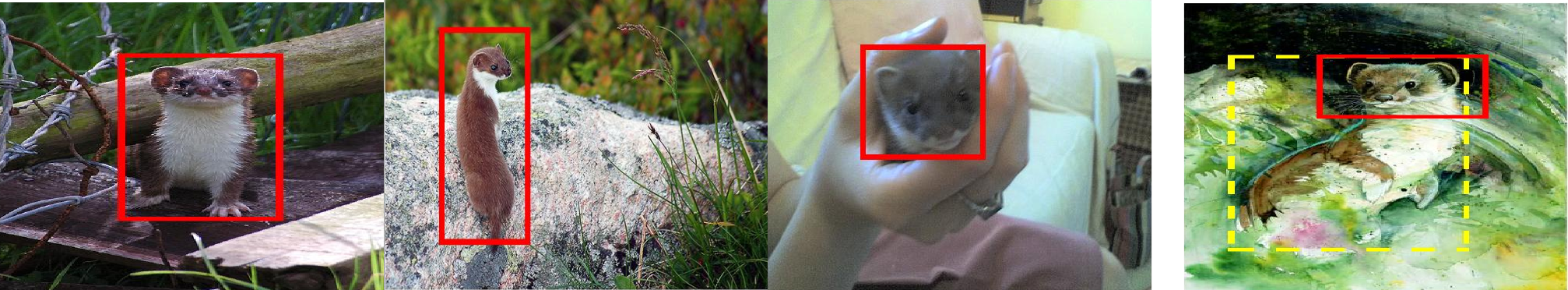} \label{fig:im3} }
 \vspace{-1em}
 \subfloat[\emph{Racoon}] { \includegraphics[width=0.65\columnwidth]{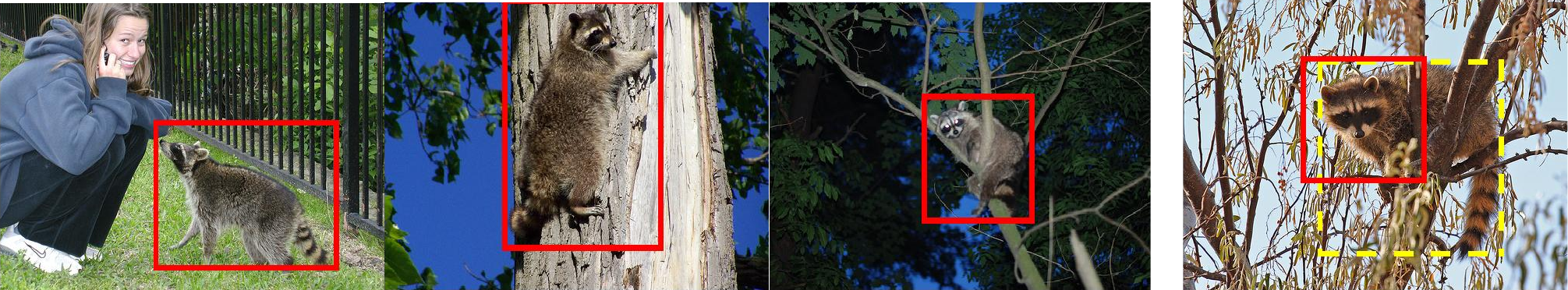} \label{fig:im4} }
	~~
 \subfloat[\emph{Rake}] { \includegraphics[width=0.65\columnwidth]{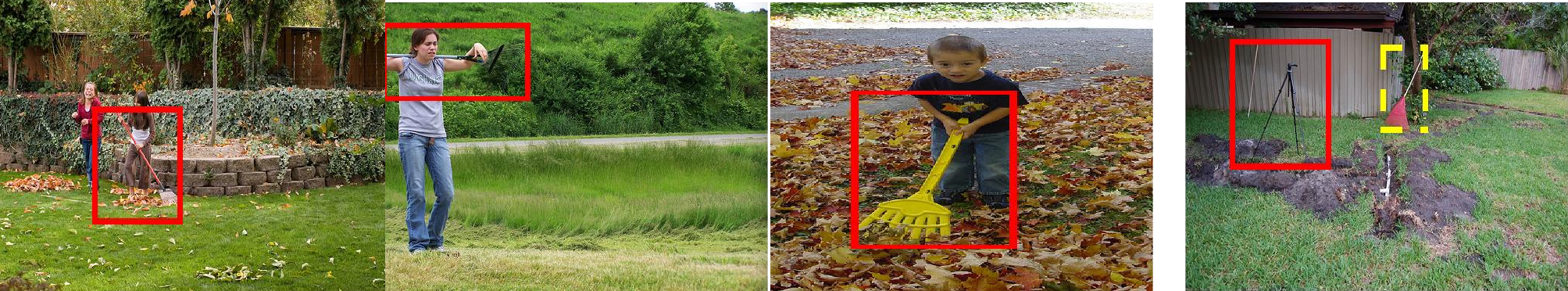} \label{fig:im5} }
	~~
 \subfloat[\emph{Wheelchair}] { \includegraphics[width=0.65\columnwidth]{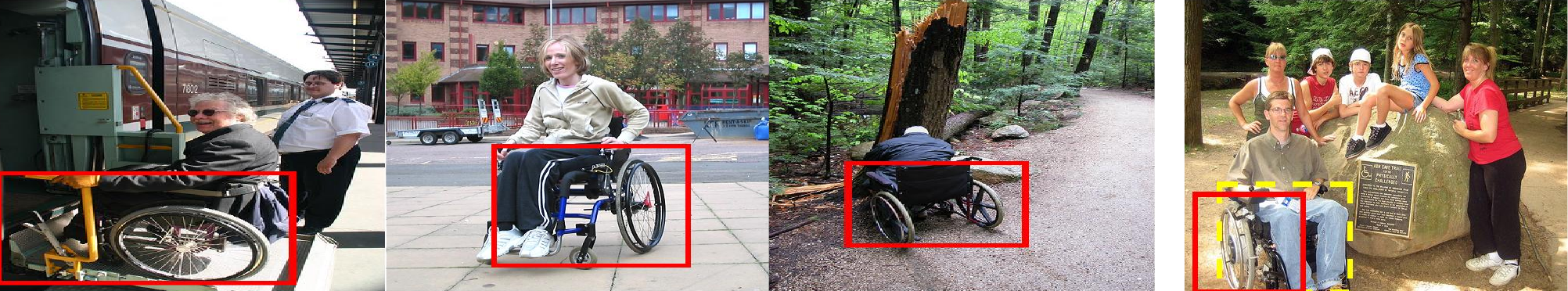} \label{fig:im6} }
 \vspace{-1em}
 \caption{Random samples of predicted object co-localization bounding box on \emph{ImageNet Subsets}. Each subfigure contains three successful predictions and one failure case. In these images, the red rectangle is the prediction by DDT, and the yellow dashed rectangle is the ground truth bounding box. In the successful predictions, the yellow rectangles are omitted since they are exactly the same as the red predictions. (Best viewed in color and zoomed in.)} \label{fig:bbox}
 \vspace{-0.7em}
\end{figure*}

Furthermore, in Fig.~\ref{fig:bbox}, several successful predictions by DDT and also some failure cases on this dataset are provided. In particular, for ``rake'' (``wheelchair''), even though a large portion of images in these two categories contain both people and rakes (wheelchairs), our DDT could still accurately locate the common object in all the images, i.e., rakes (wheelchairs), and ignore people. This observation validates the effectiveness (especially for the high CorLoc metric on \emph{rake} and \emph{wheelchair}) of our method from the qualitative perspective.

\subsection{Detecting noisy images}\label{sec:noise}

\begin{figure}[t]
 \centering
 \includegraphics[width=0.99\columnwidth]{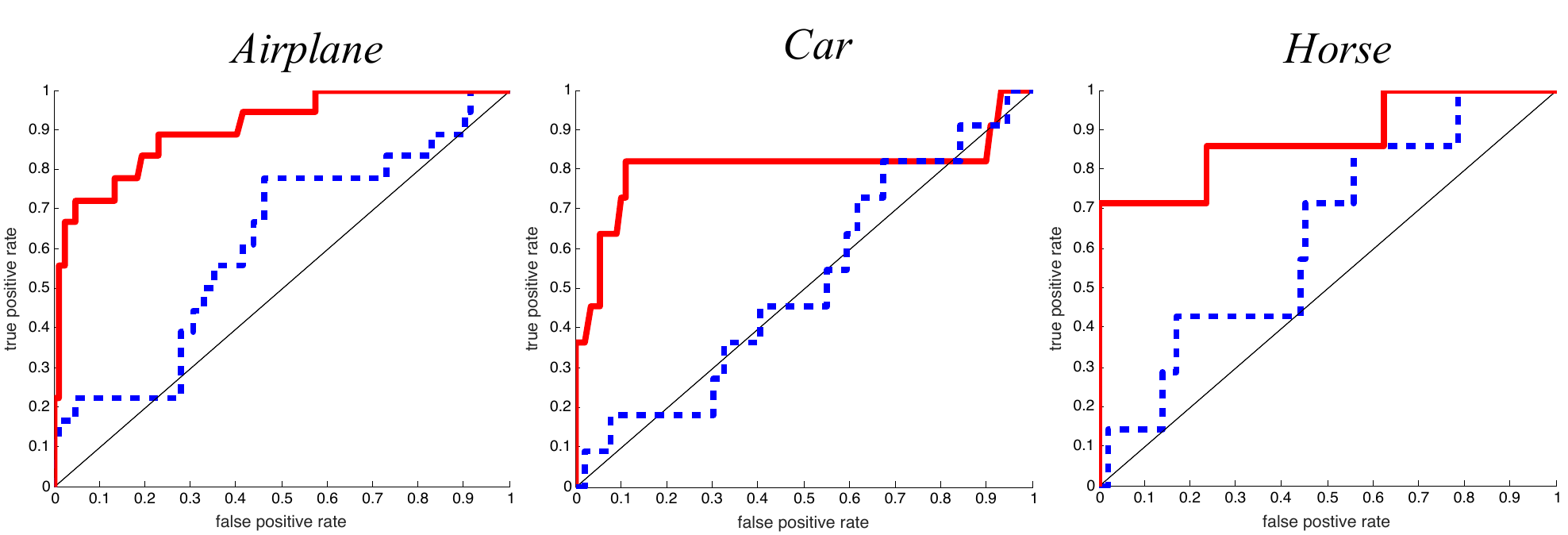}
 \vspace{-0.5em}
 \caption{ROC curves illustrating the effectiveness of our DDT at identifying noisy images on the \emph{Object Discovery} dataset. The curves in red line are the ROC curves of DDT. The curves in blue dashed line present the method in [Tang \emph{et al}., 2014].}
 \label{fig:roc}
 \vspace{-0.5em}
\end{figure}

In this section, we quantitatively present the ability of DDT to identify noisy images. As aforementioned, in \emph{Object Discovery}, there are 18\%, 11\% and 7\% noisy images in the corresponding categories. In our DDT, the number of positive values in $P^1$ can be interpreted as a detection score. The lower the number is, the higher the probability of noisy images will be. In particular, no positive value at all in $P^1$ presents the image as definitely a noisy image. For each category in that dataset, the ROC curve is shown in Fig.~\ref{fig:roc}, which measures how the methods correctly detect noisy images. In the literature, only the method in~\cite{tangcvpr2014} (i.e., the \texttt{Image-Box} model in that paper) could solve image co-localization with noisy data. From these figures, it is apparent to see that, in image co-localization, our DDT has significantly better performance in detecting noisy images than \texttt{Image-Box} (whose noisy detection results are obtained by re-running the publicly available code released by the authors). Meanwhile, our mean CorLoc metric without noise is about 12\% higher than theirs on \emph{Object Discovery}, cf. Table~\ref{table:objdisc}.

\subsection{Further study}\label{sec:future}

\begin{figure}[t]	
 \centering
 \includegraphics[width=0.9\columnwidth, height = 6.9em]{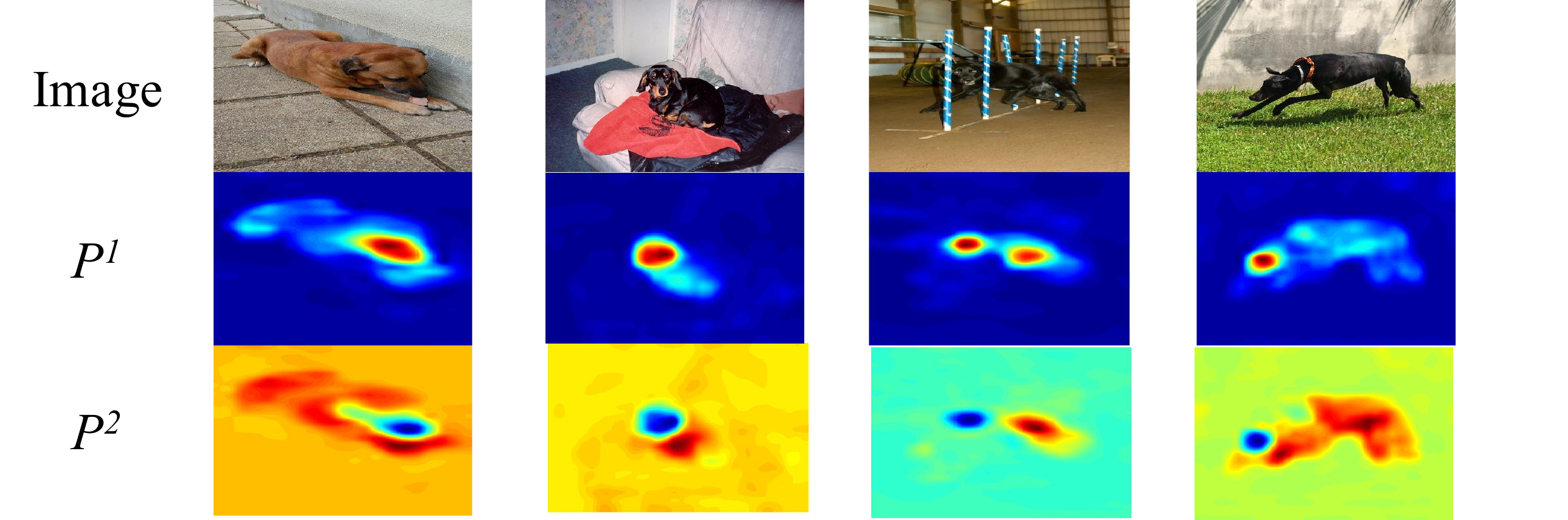}
\vspace{-0.4em}
 \caption{Four images belonging to the \emph{dog} category of \emph{VOC 2007} with visualization of their indicator matrices $P^1$ and $P^2$. In visualization figures, warm colors indicate positive values, and cool colors present negative. (Best viewed in color.)}
 \label{fig:dog}
 \vspace{-0.7em}
\end{figure}

In the above, DDT only utilizes the information of the first principal components, i.e., $P^1$. How about others, e.g., the second principal components $P^2$? In Fig.~\ref{fig:dog}, we show four images containing dogs and the visualization of their $P^1$ and $P^2$. Through these figures, it is apparently to find $P^1$ can locate the whole common object. However, $P^2$ interestingly separates the head region from the torso region. Meanwhile, these two meaningful regions can be easily distinguished from the background. These observations inspire us to use DDT for the more challenging \emph{part-based} image co-localization task in the future, which is never touched before. 

\section{Conclusions}

Pre-trained models are widely used in diverse applications in machine learning and computer vision. However, the treasures beneath pre-trained models are not exploited sufficiently. In this paper, we proposed Deep Descriptor Transforming (DDT) for image co-localization. DDT indeed revealed another reusability of deep pre-trained networks, i.e., convolutional activations/descriptors can play a role as a common object detector. It offered further understanding and insights about CNNs. Besides, our proposed DDT method is easy to implement, and it achieved great image co-localization performance. Moreover, the generalization ability and robustness of DDT ensure its effectiveness and powerful reusability in real-world applications.

DDT also has the potential ability in the applications of video-based unsupervised object discovery. In addition, robust PCA is promising to be used in DDT for improving the CorLoc metric. Furthermore, interesting observations in Sec.~\ref{sec:future} make the more challenging but intriguing part-based image co-localization problem be a future work.



\newpage
\bibliographystyle{named}
\bibliography{coloc}

\end{document}